\documentclass[10pt,twocolumn,letterpaper]{article}

\usepackage{cvpr}
\usepackage{times}
\usepackage{epsfig}
\usepackage{graphicx}
\usepackage{amsmath}
\usepackage{amssymb}

\usepackage[pagebackref=true,breaklinks=true,letterpaper=true,colorlinks,bookmarks=false]{hyperref}

\cvprfinalcopy 

\ifcvprfinal\fi
\begin{document}

\title{3D Pose Detection in Videos: Focusing on Occlusion}

\author{Justin Wang\\
{\tt\small jjwang01@stanford.edu}
\and
Edward Xu\\
{\tt\small edxu24@stanford.edu}
\and
Kangrui Xue\\
{\tt\small kangruix@stanford.edu}
\and
Łukasz Kidziński\\
{\tt\small lukasz.kidzinski@stanford.edu}
}
\maketitle

\begin{abstract}
    In this work, we build upon existing methods for occlusion-aware 3D pose detection in videos. We implement a two stage architecture that consists of the stacked hourglass network to produce 2D pose predictions, which are then inputted into a temporal convolutional network to produce 3D pose predictions. To facilitate prediction on poses with occluded joints, we introduce an intuitive generalization of the cylinder man model used to generate occlusion labels. We find that the occlusion-aware network is able to achieve a mean-per-joint-position error 5 mm less than our linear baseline model on the Human3.6M dataset. Compared to our temporal convolutional network baseline, we achieve a comparable mean-per-joint-position error of 0.1 mm less at reduced computational cost.
\end{abstract}

\section{Introduction}
Human pose detection has been an active area of research in the deep learning community since 2014 with Toshev et. al.'s DeepPose \cite{toshevdeeppose2014}, a work focused on 2D pose estimation. The problem involves detecting joint positions and bone lengths of humans in the context of both images and videos. In 3D pose estimation, additional challenges arise in regards to projecting from 2D image data onto 3D keypoint coordinates and vice-versa. However, in applications requiring extensive, realistic tracking of human motion, 3D pose estimation emerges as the only viable option. Practical examples include sports and medical fields, where athletes can analyze their footwork and body form in sharp detail, and medical professionals can deeply understand a patient's gait before entering the operating room. 

We focus on improving 3D pose estimation in video for occluded cases. Occluded joints are body joints that cannot be seen by the camera, either blocked by other joints, other body parts, or external objects. Researchers have shown that occlusion is a significant source of error in state-of-the-art models for both 2D and 3D pose estimation of single images \cite{newellhourglass2016, DBLP:journals/corr/WalkerMGH17}. A recent body of literature by Cheng et. al. \cite{chengocclusion2019} has focused on 3D pose estimations in video, specifically tackling the problem of occlusion by making use of temporal information provided by videos which is unavailable in single frames.

We build a temporal convolutional network (TCN) for 3D pose estimation to work well specifically for occlusion cases. We plan to work with well-known 3D pose datasets, HumanEva and Human 3.6M, but we will primarily focus on HumanEva due to its simplified structure and lower resource and computational power requirements. 

Since previous works have trained and fine-tuned 2D pose estimation models using Stacked Hourglass architectures on the frames of our datasets, we are able to acquire predicted 2D poses, provided by \cite{pavllovideopose2019} in the form of joint keypoints. From these 2D joint coordinates, we also generate occlusion predictions based on predefined heuristics, where we label the joint $1$ if it is occluded and $0$ otherwise. The inputs to our model are the predicted 2D joint coordinates and occlusion vector. Our labels are the 3D ground truth poses from the dataset in the form of keypoints. We also include "ground truth" occlusion labels ($0$ or $1$) generated from ground truth 3D keypoints using our own baseline heuristic. The outputs of our model are predicted 3D poses, represented as 3-dimensional coordinates in the camera's frame. 

\section{Related Work}
A landmark improvement in 2D pose estimation came via Newell et. al.'s Stacked Hourglass architecture \cite{newellhourglass2016}. Motivated by an understanding that human poses are best captured at different levels of detail (e.g. the location of faces and hands as opposed to the person's overall orientation), this architecture consists of pooling and upsampling layers which looks like an hourglass, hence the name. 

Building on the success of the Stacked Hourglass on 2D pose estimation, Martinez et. al. \cite{martinezbaseline2017} constructed a simple baseline which used the 2D pose estimations produced by the Stacked Hourglass model as inputs into a linear model to produce 3D pose estimations. 

Pavllo et. al. \cite{pavllovideopose2019} focuses on 3D pose estimation on video. Specifically, \cite{pavllovideopose2019}'s approach uses a temporal convolutional network (TCN) instead of a linear model on top of the 2D pose estimations produced by the Stacked Hourglass model. Earlier methods incorporated recurrent neural networks (RNN) to capture the temporal relationship between frames in a video, and the temporal convolutional architecture builds on this relationship by allowing for parallel processing of multiple frames, something not possible with recurrent architectures.

However, \cite{pavllovideopose2019} did not specifically focus on occlusion and therefore had many problems predicting occluded joints. A recent body of literature by Cheng et. al. \cite{chengocclusion2019} has focused on 3D pose estimations in video, specifically tackling the problem of occlusion. We seek to build off their implementation and results. To do so, we investigate why their model succeeds. Videos provide a sequence of frames that provide temporal information to better inform a model's estimations in occluded settings by providing a context in which an occluded joint should be located. Cheng et. al. \cite{chengocclusion2019} uses an occlusion-aware convolutional neural network to mitigate the effects of 3D pose estimation in occluded videos. Their "Cylinder Man Model" is a heuristic that maps 3D ground truth joint keypoints to 2D ground truth pose heatmaps and takes occlusion into account. With the 2D ground truth heatmaps, they are able to come up with occlusion labels for each of the 2D keypoints. Cheng et. al. takes predicted 2D poses from a Stacked Hourglass model, filters out occluded joints, and trains a 2D temporal convolutional network to smooth the predicted 2D keypoints. Lastly, they input the smoothed 2D keypoints and occluded predictions into a 3D temporal convolutional network and generate 3D pose predictions, using 3D ground truth poses and occlusion labels. 

Cheng et. al. primarily use 2D "ground truth" heatmaps to train a Stacked Hourglass model to output 2D predicted heatmaps with more occlusions. Although they also input occlusion labels to their TCN, we seek to actually train our TCN to recognize the ground truth occlusions by explicitly adding to the loss function. We plan to iterate upon their cylinder man model heuristic to deal with occlusion. 

\begin{figure}[t]
\begin{center}
   \includegraphics[width=0.8\linewidth]{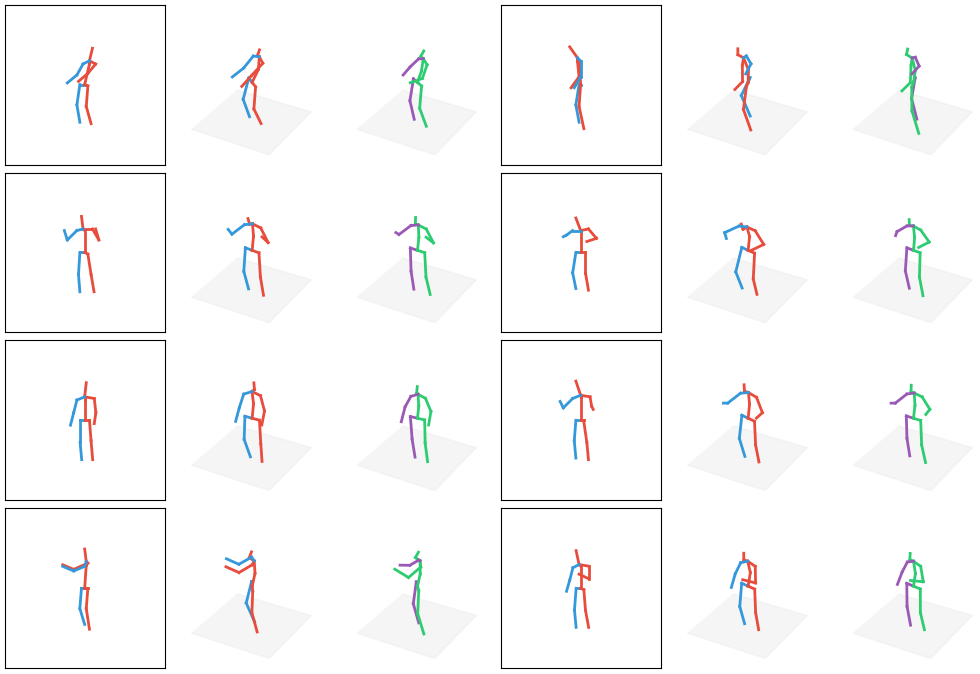}
\end{center}
   \caption{Visualization generated by Martinez et al. \cite{martinezbaseline2017}. The left column corresponds to 2D joint coordinates, the middle to ground truth 3D joint coordinates, and the right to predicted 3D join coordinates from 2D heatmaps. }
\label{fig:long}
\label{fig:onecol}
\end{figure}

\section{Datasets}

\subsection{Human 3.6M}
The Human 3.6M dataset \cite{h36m_pami}, a 3D Pose dataset, consists of 3.6 million images from actors performing a few daily-life activities. There are a total of 4 cameras and 7 annotated subjects. The preprocessed Human3.6M dataset consists of 3D ground truth joint keypoints ordered by camera used during recording, camera parameters, actions conducted by subject, and subject names. As with the temporal convolutional and linear baseline models, we use Subjects 1, 5, 6, 7, 8 for training, and reserve Subjects 9 and 11 for evaluation.  

We have access to the pre-processed dataset and have contacted the curators to gain access to the original video frames. However, we still have yet to hear a response. 

\subsection{HumanEva-I}
We have gained full access to the HumanEva-I dataset \cite{Sigal:IJCV:10b}, which contains 3D ground truth keypoints per frame, representing 15 joints (pelvis, thorax, shoulders, elbows, wrist, hips, knees, ankles, head), along with their original video frames. There are a total of seven video-sequences (four grayscale and three color) of four annotated subjects performing five different actions (Walking, Jogging, Throwing/Catching, Gesturing, and Boxing).

We generate our training, validation, and test data using a modified version of the pre-processing step used in \cite{pavllovideopose2019}. Specifically, this pre-processing step involves calculating the projection from 3D ground truth joint keypoints to 2D joint keypoints using the provided motion capture and camera calibration data. Additionally, since the original video frames occasionally contain chunks of invalid joint keypoint measurements, we simply discard these. 

For the remaining video frames, we only consider the three color cameras, the first three subjects, and the first video sequence take of each action. This leaves us with a total of 28,731 data entries, partitioned into a roughly 50/50 training and validation split.

\section{Method}

\subsection{Clustered Ground Truth Occlusions}
In order to generate "ground truth" occlusion labels for 2D poses, we start with a simple heuristic called "Clustered Occlusions", similar to the method of \cite{chengocclusion2019}. For every frame, we have 17 joint keypoints $k$ in the form of 3D camera coordinates $k_i=(x_i, y_i, z_i)$ for the Human3.6M dataset, or 15 joint keypoints for the HumanEva-I dataset. Because most of the joints in these frames are occluded by other body parts or joints, our intuition to finding which joints are occluded is to find joints that are clustered together in the $xy$-plane, then mark the joint closest to the camera (smallest depth) as not occluded, and mark the other joints in the cluster as occluded. In other words, for each joint coordinate $k_i$, we find the set of keypoints $k_j$ where
\[\sqrt{(x_i-x_j)^2+(y_i-y_j)^2}<\epsilon\]
and $\epsilon$ is the tunable tolerance parameter. We currently use $\epsilon=0.06$. Then, we add $k_i$ and all keypoints $k_j$ to form a cluster $S$, and our non-occluded keypoint index $n$ from this set $S$ is 
\[n=\arg\min(z_p)\]
where $k_p\in S$. We mark all other points in set $S$ as occluded. We hope that this heuristic can generally fetch all joints that are observable by the camera, as we believe joints in close proximity occlude one another. This gives us a vector of occluded joints $o$ for each frame, where $1$ means occluded and $0$ means not occluded.  

After applying the heuristic to get occluded joints, we generate ground truth 2D heatmaps for each existing 2D pose by placing a white circle with Gaussian smoothness at the image coordinates for joints that are not occluded, and doing nothing for joints that are occluded. Then, we take a center crop of the heatmap to crop the subject and resize the width and height to 128. Our ground truth and predicted 2D heatmaps can be visualized in Figures 2 and 3. 

\begin{figure}[t]
\begin{center}
   \includegraphics[width=0.8\linewidth]{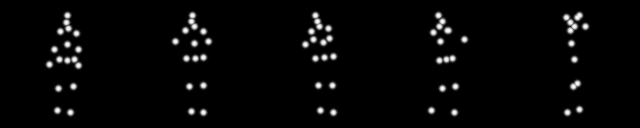}
   \includegraphics[width=0.8\linewidth]{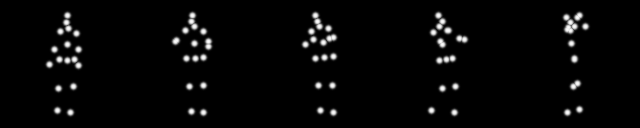}
\end{center}
   \caption{Ground truth 2D heatmaps (top) and predicted 2D heatmaps (bottom) for a sequence of poses. The ground truth heatmaps have less peaks (occluded keypoints).}
\label{fig:long}
\label{fig:onecol}
\end{figure}

This is only a simple heuristic, and we hope to test out how adding the ground truth occlusion labels affect our error. 

\begin{figure}[t]
\begin{center}
   \includegraphics[width=0.45\linewidth]{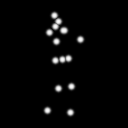}
   \includegraphics[width=0.45\linewidth]{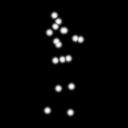}
\end{center}
   \caption{Up close example of the occluded joints omitted in ground truth heatmap (top) compared to predicted heatmap (bottom).}
\label{fig:onecol}
\end{figure}

\subsection{Boxed Man Model}
To improve our heuristic, we take inspiration from the cylinder man model delineated in \cite{chengocclusion2019}. The cylinder man model generates occlusion labels for 2D poses using 3D poses (Figure 3). Specifically, it models the head, body, arms, and legs as cylindrical segments. In general, if a certain joint is located within another joint's cylindrical segment in 3D space, it is deemed occluded in the 2D space. More specifically, this technique calculates a visibility variable to determine the degree of occlusion for each joint.

We take this idea and adapt it to our computing needs, and introduce an occlusion technique that requires less computational power and memory than the original cylinder man model. We propose a boxed man model that generates occlusion labels for 2D poses using the original 2D poses. We visualize the original cylinder man, with equivalent proportions, squashed into 2D space.

For example, in the case of the Human3.6M dataset, keypoints 9 and 10 represent the top and bottom of the subject's head. Define these keypoints to be $A$ and $B$, as seen in Figure 4. We use these two keypoints and project them to four points $A_1, A_2, B_1, B_2$ which determine the bounds of our boxed approximation of the head. We determine the four points by first calculating slope of the line $AB$, which we define as $m_{AB}$, to then find the perpendicular slope $m'_{AB} = -\frac{1}{m_{AB}}$. We then define
\begin{align*}
    A_1 &= (A_x - \cos(\arctan(m'_{AB})) * \delta,\\ &A_y - \sin(\arctan(m'_{AB}))) * \delta)\\
    A_2 &= (A_x + \cos(\arctan(m'_{AB})) * \delta,\\ &A_y + \sin(\arctan(m'_{AB}))) * \delta)
\end{align*} 
where $\delta$ is a hyperparameter that determines the width of the boxes. $B_1$ and $B_2$ are defined similarly and instead use $B$'s coordinates $B_x$ and $B_y$. The box that entails the chest and torso area is defined by the four points that are provided in the keypoints, specifically 1, 4, 11, 14 in the Human3.6M dataset.

If a joint in the boxed man model is located within another joint's boxed segment in 2D space, we deem it occluded. This simpler heuristic encourages our temporal convolutional network to learn poses based on joints which definitively not occluded. We believe that the temporal convolutional network will be able to learn poses for occluded joints from other camera positions in the dataset, and should learn as much as possible from the original data rather than be forced to learn from certain joints by a heuristic.

\begin{figure}[t]
\begin{center}
   \includegraphics[width=0.8\linewidth]{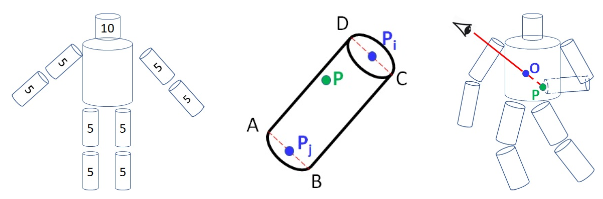}
\end{center}
   \caption{A visualization of the cylinder man model \cite{chengocclusion2019}. Each arm and leg are scaled to a diameter of 5 cm, and the head is scaled to a diameter of 10 cm.}
\label{fig:my_label}
\end{figure}

\begin{figure}[t]
\begin{center}
   \includegraphics[width=0.8\linewidth]{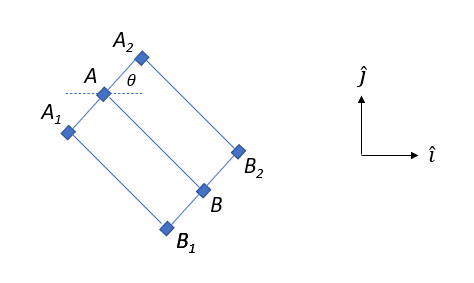}
\end{center}
   \caption{The setup for the boxed man model. Note that $\hat{i}$ and $\hat{j}$ represent unit vectors for the $x$ and $y$ axes.}
\label{fig:my_label}
\end{figure}

\subsection{Temporal Convolutional Network Model}
Our main model that produces 3D poses is the 3D temporal convolutional network (TCN) adapted from Pavllo et. al. \cite{pavllovideopose2019}. Taking a consecutive sequence of 2D joint keypoints, it uses temporal convolutions and residual connections to predict a frame's 3D joint coordinates. The input layer applies a temporal convolution over the 2D keypoints with kernel size $W=3$, expanding the number of channels from two times the number of joints (for each $x$ and $y$ coordinate) to $C=1024$. Then, the model goes through $B$ Resnet-type blocks, which are connected through skip layers. Each block has a 1D convolution with kernel size $W$ and $C$ channels, followed by another convolution with kernel size $1$. All convolutional layers are followed by batch normalization, ReLU activation, and dropout layers. Finally, the last layer shrinks the number of channels and outputs the predicted 3D pose keypoints. 

Because we are focusing on occlusion, we also input a sequence of predicted occluded joints into our TCN, where every joint is either $0$ for not occluded or $1$ for occluded. First, we apply a temporal convolution to the occluded vectors. Then, we apply sigmoid activation and zero out keypoints whose values in the occluded vectors are above some threshold. By doing this, we are effectively trying to learn the ground truth occluded vectors over the temporal convolution, hoping that we can successfully zero out the actually occluded keypoints before applying convolutions over the 2D keypoint sequence. 

We explore two variants of our model with this input. Our first variant uses down-convolutions to drop from a temporal range of vectors to one occluded vector for the frame of the 3D pose we are predicting. This way, the outputted occluded vector can be directly compared to the frame's ground truth occluded vector in the loss. Instead of focusing on learning only one occluded vector, our second variant does not have any down-convolutions and tries to learn the whole sequence of ground truth occluded vectors.

\subsection{Loss Function}
With the ground truth occlusion labels, we can now train our TCN to notice the occluded joints. To do so, we can add a loss to our existing loss function $L$, which is the mean per joint positional error (MPJPE) between estimated and ground truth 3D poses:
\[L=\frac{1}{M} \frac{1}{N} \sum_{k=1}^M \sum_{i=1}^N ||(J_i^{(k)} - J_{root}^{(k)}) - (\hat{J}_i^{(k)} - \hat{J}_{root}^{(k)})||_2\]
where $M$ represents the number of examples and $N$ is the number of joints. Now, let the ground truth occlusion vector for a frame be $o$ and the predicted occlusions be $\tilde{o}$, our modified loss function $L'$ will be:
\[L'=\lambda_1L+\lambda_2\frac{1}{M} \frac{1}{N} \sum_{k=1}^M \sum_{i=1}^N|o_i^{(k)}-\tilde{o}_i^{(k)}|\]
where $\lambda_1$ and $\lambda_2$ are tunable weights. 


\section{Experiments}

\subsection{Baselines}
Pavllo et. al. \cite{pavllovideopose2019} implemented 3D pose estimation in videos using temporal convolutions and semi-supervised training. We use this model as a baseline for our 3D pose estimation as the model does not actively seek to solve the problem of occlusion. 

Another baseline whose work we seek to build upon and compare against is \cite{chengocclusion2019}. Because their network is occlusion-aware, we hope to achieve similar results or possibly improve on the problems that they face.  

\subsection{Results}
We evaluate our models using the original mean per joint positional error (MPJPE) between estimated and ground truth 3D poses in millimeters (mm). The mean is calculated over all $N$ joints used in each image frame; in this case, $N = 17$ for the Human3.6M dataset, and $N = 15$ for the HumanEvaI dataset. We first align the pelvis as the root joint before comparing differences in Euclidian distance between the poses. The joints are also normalized with respect to the root joint.

\subsection{Results on HumanEva-I}
We trained and evaluated our model on HumanEva, looking at Subjects 1, 2, and 3 and Actions Walk, Jog, and Box as they are the subjects and actions focused on in \cite{pavllovideopose2019}. We seek to compare mostly against \cite{pavllovideopose2019} because we use the same network, but we also compare our results to \cite{chengocclusion2019}. Results are shown in Tables 1 through 5. Tables 1 through 3 show the results for different methods over actions, Table 4 shows Cheng et.al.'s results, and Table 5 shows the averages over all actions and subjects for methods. 

\begin{table}[h!]
\centering
 \begin{tabular}{||c||c c c||c||} 
 \hline
 Walk & Subject 1 & Subject 2 & Subject 3 & Average \\ [0.5ex] 
 \hline\hline
 1 & 13.9 & 10.2 & 46.6 & 23.6 \\
 2 & 14.4 & 10.2 & 46.8 & 23.8 \\
 3 & 14.1 & \textbf{10.0} & 46.7 & 23.6 \\
 4 & \textbf{13.8} & 10.1 & \textbf{46.5} &  \textbf{23.5} \\
 \hline
 \end{tabular}
\caption{MPJPE of the action Walking for HumanEva (mm). (1: Pavllo et.al. \cite{pavllovideopose2019}; 2: One vector, Clustered; 3: Many vectors, Clustered; 4: Many vectors, Boxed Man). Our multiple occlusion vector method coupled with the boxed man model keypoints achieves the best average MPJPE across subjects. Bolded numbers are the best among the methods. }
\label{table:1}
\end{table}

\begin{table}[h!]
\centering
 \begin{tabular}{||c||c c c||c||} 
 \hline
 Jog & Subject 1 & Subject 2 & Subject 3 & Average \\ [0.5ex] 
 \hline\hline
 1 & 20.9 & 13.1 & 13.8 & 15.9 \\
 2 & \textbf{20.7} & \textbf{13.0} & 13.7 & \textbf{15.8} \\
 3 & 21.0 & 13.1 & 13.8 & 16.0 \\
 4 & 21.1 & 13.0 & \textbf{13.7} & 15.9 \\
 \hline
 \end{tabular}
\caption{MPJPE of the action Jogging for HumanEva (mm). (1: Pavllo et.al. \cite{pavllovideopose2019}; 2: One vector, Clustered; 3: Many vectors, Clustered; 4: Many vectors, Boxed Man). Our one vector method coupled with the simple clustered heuristic achieves the best average MPJPE across subjects. Bolded numbers are the best among the methods. }
\label{table:2}
\end{table}

\begin{table}[h!]
\centering
 \begin{tabular}{||c||c c c||c||} 
 \hline
 Box & Subject 1 & Subject 2 & Subject 3 & Average \\ [0.5ex] 
 \hline\hline
 1 & 23.8 & 33.7 & 32.0 & 29.8 \\
 2 & \textbf{23.7} & \textbf{33.0} & 32.0 & \textbf{29.6} \\
 3 & 23.9 & 33.2 & 31.7 & 29.6 \\
 4 & 23.9 & 33.4 & \textbf{31.6} & 29.6 \\
 \hline
 \end{tabular}
\caption{MPJPE of the action Boxing for HumanEva (mm). (1: Pavllo et.al. \cite{pavllovideopose2019}; 2: One vector, Clustered; 3: Many vectors, Clustered; 4: Many vectors, Boxed Man). Our one vector method coupled with the simple clustered heuristic achieves the best average MPJPE across subjects. Bolded numbers are the best among the methods. }
\label{table:3}
\end{table}

\begin{table}[h!]
\centering
 \begin{tabular}{||c||c c c||c||} 
 \hline
 Action & Subject 1 & Subject 2 & Subject 3 & Average \\ [0.5ex] 
 \hline\hline
 Walk & 11.7 & 10.1 & 22.8 & 14.9 \\
 Jog & 18.7 & 11.4 & 11.0 & 13.7 \\
 \hline
 \end{tabular}
\caption{MPJPE for HumanEva by Cheng et.al. \cite{chengocclusion2019} (mm). They seem to have amazing performances across the board, most likely because they have extend their method from end-to-end and use heatmaps to be occlusion-aware. We do not have the compute power that they do to add heatmaps to our model.}
\label{table:4}
\end{table}

\begin{table}[h!]
\centering
 \begin{tabular}{||c||c||} 
 \hline
 Method & Average \\ [0.5ex] 
 \hline\hline
 Pavllo et.al. \cite{pavllovideopose2019} & 23.11 \\
 One vector, Clustered & 23.06 \\
 Many vectors, Clustered & 23.06 \\
 Many vectors, Boxed Man & \textbf{23.01} \\
 \hline
 \end{tabular}
\caption{Average MPJPE for HumanEva over all subjects and actions considered (mm). Our many vector variant with the boxed man model occlusion keypoints seems to work the best, beating the baseline and our other methods. }
\label{table:5}
\end{table}

From these results, our variant of the temporal convolutional model that compares a sequence of occluded vectors to the sequence of ground truth vectors, coupled with our Boxed Man Model, works the best, achieving an average MPJPE of 23.01 over all subjects and actions. Our other variant that only uses the single frame occluded vector in the loss also performed better than our initial baseline, showing that the Clustered occlusion heuristic also worked well to prevent occluded joints from being wildly predicted. 

\subsection{Baseline Results on Human3.6M}
After one epoch and convergence respectively, the linear baseline was able to achieve the following results on 3 of the 15 tasks, and on average:
\begin{table}[h!]
\centering
 \begin{tabular}{||c c c||c||} 
 \hline
 Directions & Photo & SittingDown & Average \\ [0.5ex] 
 \hline\hline
 60.86 & 95.53 & 117.99 & 77.24\\ 
 39.5 & 56 & 69.4 & 47.7 \\
 \hline
 \end{tabular}
\caption{MPJPE of the linear baseline (mm).}
\label{table:6}
\end{table}

After convergence, as reported in \cite{chengocclusion2019}, the occlusion aware network was able to achieve the following results:
\begin{table}[h!]
\centering
 \begin{tabular}{||c c c||c||} 
 \hline
 Directions & Photo & SittingDown & Average \\ [0.5ex] 
 \hline\hline
 38.8 & 51.9 & 58.4 & 42.9 \\
 \hline
 \end{tabular}
\caption{MPJPE of the occlusion-aware model (mm).}
\label{table:7}
\end{table}

\subsection{Results on Human3.6M}
We initially tried inputting 32 by 32 predicted and ground truth heatmaps into the TCN to combat occlusion, similar to \cite{chengocclusion2019}. However, because of our lack of computational power, we only were able to train a subset of the data for a few epochs. We ended up with an test error of 174.66 mm. We also tried changing from occluded heatmaps to occluded vectors, and we ended up with a test error of 50.36 mm. Our results on Human 3.6M are definitely stunted by the low computation power and the sheer size of the dataset. 

\subsection{Discussion}
\subsubsection{Baseline}
For our baseline results, we selected the three tasks Directions, Photo, and SittingDown, because we believe they best exhibit the differences in the model's performances. The difference in performance on the Photo action is around the mean of the differences in performance with respect to all actions. While the variation in MPJPE for the Directions task is small between the linear baseline and occlusion aware network, it is significantly larger for the SittingDown action. We believe this is due to the occluded nature of the action, as well as the significance of temporal information. Sitting down involves the knee joints occluding the hip joint at the end of the action from a ventral camera orientation. Given that the linear model is trained over single images, it is unable to learn the hip joint's trajectory over the course of multiple frames. On the other hand, the occlusion aware network is able to use both its heuristic of whether or not the hip joint is occluded along with the hip joint's trajectory via a sequence of video frames to predict where the hip joint should be located accurately.

\subsubsection{Boxed Man Model}
Naturally, due to the boxed man model's anatomically derived occlusion method, it performed better than the baseline Gaussian model. We believe that another source of increased performance for the boxed man model is its stronger tendency to mark a given joint as occluded as opposed to the baseline. Occlusion serves as a form of dropout or regularization of the model. Specifically, we believe that feeding the network information about whether or not a joint is occluded eventually teaches the model to rely less on joints that are marked as occluded. Given that different actions cause inherently different occlusion patterns, the model will be less inclined to focus on a few joints during training. The boxed man model's lax requirements on occlusion allows for a wider variation of non-occluded keypoint permutations.

\section{Conclusion}

Most of our work focuses on adapting temporal convolutional network to predict occluded human 3D poses from 2D ground truth heatmaps, and indeed, the average mean-per-join-position error across our three network variants was comparable to but nonetheless still lower than the baseline architecture in \cite{pavllovideopose2019}. However, since this only constitutes the second half of the video to estimated 3D human pose pipeline, work remains to be done in improving occluded heatmap generation in the first place. 

For example, training a stacked hourglass network using our clustered ground truth occlusions and boxed man models would make our task more consistent end-to-end. We initially began experimenting with such methodologies and re-configured an existing stacked hourglass implementation \cite{newellhourglass2016} (originally configured to work with the MPII human pose dataset) to work with the HumanEva-I dataset. However, due to limited computing resource, we decided to focus on training the temporal convolutional network.

Another similar idea involves adding data augmentation by manually blocking out non-occluded joints. This could simply take the form of dropping out random joints during training of the stacked hourglass network, or since significant pre-processing is already being performed on the videos, it could perhaps even involve editing the frames themselves.

{\small
\bibliographystyle{ieee}

}

\end{document}